%% file: main.tex
\documentclass[sigconf,nonacm,9pt]{acmart}

\usepackage{booktabs} 
\usepackage{styles/all}

\begin{document}
\title{Improving Video Compression With Deep Visual-Attention Models}

\author{Vitaliy Lyudvichenko}
\affiliation{\institution{Lomonosov Moscow State University}}
\email{vlyudvichenko@graphics.cs.msu.ru}

\author{Mikhail Erofeev}
\affiliation{\institution{Lomonosov Moscow State University}}
\email{merofeev@graphics.cs.msu.ru}

\author{Alexander Ploshkin}
\affiliation{\institution{Lomonosov Moscow State University}}
\email{alexander.ploshkin@graphics.cs.msu.ru}

\author{Dmitriy Vatolin}
\affiliation{\institution{Lomonosov Moscow State University}}
\email{dmitriy@graphics.cs.msu.ru}

\begin{abstract}
Recent advances in deep learning have markedly improved the quality of visual-attention modelling. In this work we apply these advances to video compression.

We propose a compression method that uses a saliency model to adaptively compress frame areas in accordance with their predicted saliency.
We selected three state-of-the-art saliency models, adapted them for video compression and analyzed their results.
The analysis includes objective evaluation of the models as well as objective and subjective evaluation of the compressed videos.

Our method, which is based on the x264 video codec, can produce videos with the same visual quality as regular x264, but it reduces the bitrate by 25\% according to the objective evaluation and by 17\% according to the subjective one.
Also, both the subjective and objective evaluations demonstrate that saliency models can compete with gaze maps for a single observer.

Our method can extend to most video bitstream formats and can improve video compression quality without requiring a switch to a new video encoding standard.

\end{abstract}


 \thanks{This work was partially supported by Russian Foundation for Basic Research under Grant 19-01-00785~a.}

\maketitle

\section{Introduction}
\label{sec:intro}

The share of video in overall Internet traffic is growing rapidly. According to the Cisco Visual Networking Index Forecast~\cite{cisco-report}, video traffic will constitute 82 percent of all IP traffic by 2022, up from 75 percent in 2017. This projection indicates an urgent need to improve the industry's video compression techniques, because even a minor improvement in a widely used video encoder will considerably reduce video traffic globally.

Numerous efforts to develop new video-encoding standards are under way (e.g., AV1 and VVC). Although adoption of these standards promises much better compression, the process will require that viewing devices be updated and will take several years.

Another possible direction for boosting the video compression ratio while maintaining the same or better visual quality is to develop better encoders that are compatible with widely adopted standards.
The bitstream format may limit the extent of such improvements, but these encoders can enter service immediately since no device update is necessary.

This paper explores the second option: build a better encoder for existing standard. In particular we focus on the relatively underused (at least by industry) approach of saliency-aware compression.

This paper explores the second option: build a better encoder for existing standard. In particular we focus on the relatively underused (at least by industry) approach of saliency-aware compression. It has long been known that visual attention is unevenly distributed across images~\cite{Yarbus1967}: whereas certain salient parts of an image may receive the most gazes, nonsalient regions may be virtually invisible to a naive observer. The natural way to exploit this feature of the human visual system is to use fewer bits to encode nonsalient regions and thereby reduce the transmission and storage costs for those regions.

We propose an H.264-compatible saliency-aware video encoder with a built-in visual-attention model (see Section~\ref{sec:proposed} for a description of the encoder structure). Our research builds on the modified x264 software encoder in~\cite{SAVAM2}. This encoder distributes the bitrate according to the sequence of per-frame saliency maps, which are given as an additional input. Achieving the bitrate redistribution involves tweaking the encoder's internal structure: macroblock-delta-quantization parameters. Notably, the proposed encoder can be reimplemented for other video-encoding standards, since most modern ones (e.g. VP9~\cite{vp9-standard} and HEVC~\cite{hevc-standard}) work with macroblock delta quantizers.

For visual-attention prediction we employ state-of-the-art saliency models. Despite good progress in saliency prediction thanks to deep learning, research into the applicability of novel visual-attention models to real world cases is lacking.

The main contribution of this paper is our thorough analysis of saliency models applicability to video compression. In both the objective and subjective evaluations (see Section~\ref{sec:CompressionSubjectiveEvaluation}) we show that using these models in our proposed encoder can reduce the bitrate by up to 17\% without sacrificing visual quality relative to naive saliency-unaware encoding. This result leads us to conclude that modern deep saliency models are good enough for the industry to adopt for real-world video compression.

\section{Related work}
\label{sec:related}

Our proposed video-compression method aims to combine recent advances in two fields: visual-attention modelling with neural networks and saliency-aware video compression. The following discussion reviews related works from both fields.

\subsection{Visual-Attention Modelling}

\textbf{Non-deep models.} 
Before the deep-learning era, saliency models mainly employed a bottom-up approach and relied on handcrafted features.
Consult~\cite{Borji2013} for a comprehensive overview of non-deep models.
But the quality of these models limited their applicability to saliency-aware compression~\cite{Gitman2014,SAVAM2}.
In particular, Judd et al.~\cite{Judd2012} showed that most of them cannot compete with a simple center-prior model.

\begin{sloppypar}
\textbf{Static models.} 
An early deep saliency model~\cite{Theis2015} used the sum of the center prior and the output from AlexNet's convolution layers to predict saliency; it outperformed all other models on the MIT benchmark. 
\end{sloppypar}

Kruthiventi et al.~\cite{Kruthiventi2017} proposed a fully convolutional network based on VGG-16. They increased the receptive field by applying dilated convolutions to capture the global context. To overcome the inability of fully convolutional networks to learn location-dependent patterns, their approach feeds into  the network a series of precomputed images with center prior.

A more recent  effort~\cite{SAM-cornia2018} introduced an advanced network architecture that incorporates an attentive model with a convolutional LSTM layer to iteratively update the network's attention. This architecture joins the output of the attentive module with 16 trainable center-prior images to produce the final saliency map. Also, it uses a combined loss function that is a linear combination of three classical saliency metrics. The abovementioned improvements helped to achieve state-of-the-art results on the SALICON data set according to several metrics.

\textbf{Dynamic models.} 
Unlike static models, dynamic models may benefit from using temporal cues (e.g., object motion). Only a few deep saliency models for video have been proposed, however.

Bak et al.~\cite{Bak2018} employed a two-branch network where the first branch extracts motion features from externally computed optical-flow maps and the second extracts image features.
Next, they applied a series of convolutional layers to the fused feature maps followed by the final deconvolution layer.
Jian et al.~\cite{OM-CNN-Jiang2017} also used a two-branch architecture.
They computed motion features internally using a subnet of FlowNet.
Then they joined the motion and image features and fed them to two sequential convolutional LSTM layers followed by the final convolutional layers.

Wang et al.~\cite{ACL-wang2018} incorporated an attention module into a network that predicts static saliency. 
Their network performs element-wise multiplication of the last feature map from  modified VGG-16 by the static saliency and passes the result to a convolutional LSTM layer.
To extend the receptive field, the authors removed some max-pooling layers from VGG and used dilated convolutions.
The training uses a combined loss function, as in~\cite{SAM-cornia2018}.
The attention module allows the network to be trained on both image and video saliency datasets.

For a more detailed overview of existing deep saliency models, consult ~\cite{Borji2018}.

\subsection{Saliency-Aware Video Compression}
The methods designed to spatially distribute bitrate according to a visual-attention map are divisible into two categories: implicit and explicit.

Implicit methods apply preprocessing to the input signal before feeding it into a saliency-unaware encoder. The authors of~\cite{Itti2004,polakovivc2018approach} apply nonuniform blur to the input frame to force the desired bit allocation. They show that preprocessed videos exhibit better visual quality than unprocessed ones encoded at the same bitrate.

A remarkable approach proposed by Zund et al.~\cite{zund2013content}. The authors apply nonuniform scaling to an input image in accordance with a saliency map (image retargeting) and then compute the difference between the source image and the downscaled/encoded image upscaled to its original size. The method transmits to the decoder a downscaled image, the grid coordinates required to unwarp it, and the difference image. Both images are compressed using a conventional method. This method requires changes to the decoder and thus precludes implementation without violating current video- and image-encoding standards.

Explicit methods deeply integrate with encoders to modify their internal data structures and algorithms for bit allocation. The authors of~\cite{gupta2011scheme} modify the macroblock-quantizer map in accordance with its mean saliency. Hadizadeh et al.~\cite{hadizadeh2014saliency}, besides tweaking macroblock quantizers, also add a new term for the rate-distortion cost function that penalizes any change in macroblock saliency due to introduced compression artifacts.

The method proposed in~\cite{zhu2018spatiotemporal} adds visual-attention awareness to the HEVC reference encoder. Instead of setting an exact quantizer for each macroblock according to its saliency, the authors propose setting ranges in which the rate-control subsystem is free to choose a particular quantizer for the given macroblock. Furthermore, they change the rate-distortion cost function by weighting its cost term with the block's saliency. The paper is also the first attempt to employ saliency maps generated using a deep-neural-network model for video compression. Our paper improves the results by analyzing how the choice and tuning of the deep visual-attention model affects compression quality.

Lyudvichenko et al.~\cite{SAVAM2} proposed saliency-aware modification of the x264 encoder. This modified encoder selects macroblock quantizers in order to use $b$ percent of the bitrate to encode $p$ percent of the least-salient pixels, where $b$ and $p$ are user defined. It is an element of our proposed method owing to the prevalence of the H.264 standard, the availability of the modified source code~\cite{sa-x264} and the x264 encoder's maturity.

\section{Proposed Video Compression Method}
\label{sec:proposed}

\input{figures/encoder_schema.tex}

The proposed saliency-aware encoder comprises two parts: the saliency-prediction module and the encoding module (see Figure~\ref{fig:encoder-schema}). It first feeds the input video into the saliency-prediction module, then it feeds the predicted saliency maps along with the input video into the encoding module, which outputs the encoded bitstream.

We use the saliency-aware modification of the x264 software encoder proposed in~\cite{SAVAM2} as our encoding module. This modified encoder chooses macroblock delta quantizers such that $b$ percent of the bitrate encodes $p$ percent of the least-salient pixels in accordance with the input saliency map. The parameters $p$ and $b$ are user defined; we fixed them at 80\% and 70\%, respectively, in all our experiments. To achieve the abovementioned bitrate distribution, our approach used the following update rule:

\newcommand{\SM}{\mathbf{S}}
\newcommand{\SP}{\mathbf{SP}}
\newcommand{\SN}{\mathbf{SN}}
\newcommand{\Bi}{\mathbf{B}}
\newcommand{\Qu}{\mathbf{Q}}

\begin{equation}\label{eq:alphabeta}
    \left\{
    \begin{aligned}
        \Qu' &= \Qu + \alpha\SP - \beta\SN	         				\\
	    \sum_{i,j: \SN   >  0} \Bi(\Qu_{i,j}') &=
			    \frac{b}{100} \sum_{i,j} \Bi(\Qu_{i,j})         	\\
    \sum_{i,j: \SP \ge 0} \Bi(\Qu_{i,j}')  &=
        \left(1 - \frac{b}{100}\right) \sum_{i,j} \Bi(\Qu_{i,j}), 	\\
    \end{aligned}
    \right.
\end{equation}
where $\Qu$ and $\Qu'$ are the original and modified quantization maps, respectively; $\Bi(q)$ estimates the number of bits required to encode a block with quantizer $q$; $\SP = \max(S - s_p, 0)$; and $\SN = \max(s_p - S, 0)$. $S$ is the input saliency map and $s_p$ is the value of its $p$-th percentile.

For the saliency prediction module we considered three state-of-the-art saliency models (see Section~\ref{subsec:models-selection}). We fine-tuned each model (see Section~\ref{subsec:models-preparation}) to estimate its generalization ability by comparing the performance of the original and fine-tuned models for our dataset.

We compared our proposed encoder with the unmodified x264 encoder both objectively (see Section~\ref{sec:CompressionObjectiveEvaluation}) and subjectively (see Section~\ref{sec:CompressionSubjectiveEvaluation}).

\section{Experiments}
\label{sec:experiments}

\subsection{Saliency Models Selection}
\label{subsec:models-selection}

We selected three state-of-the-art visual-attention models as candidates for the saliency-prediction module. Two of them, OM-CNN~\cite{OM-CNN-Jiang2017} and ACL~\cite{ACL-wang2018}, are designed for saliency prediction in videos, and the third one, SAM~\cite{SAM-cornia2018} (we use its ResNet version), is for images.

\input{figures/models_examples.tex}

SAM is unable to exploit motion features, but it strongly exploits image structural features because it was trained on a more diverse image dataset containing thousands of images. OM-CNN, on the other hand, was trained on datasets with only dozens of videos. Therefore, static visual-attention models may be better at generalization and may produce better saliency maps for videos with rare or previously unseen types of content. ACL combines both approaches; its network has an attention module trained on both images and videos.

We also compare these models using three baselines: center prior, single observer, and ground truth. Figure~\ref{fig:ModelExamples} shows example saliency maps for the models and the baselines.

These baselines have a meaningful value for sali\-en\-cy-aware-compression research:
\begin{itemize}
\item Though center prior is a simple static model that does not depend on video content, it outperforms some non-deep models~\cite{SAVAM2}.

\item Single-observer saliency maps are based on eye-tracking data from one observer. Until recently, automatic models had lower quality than the single-observer baseline~\cite{Judd2012}. Also, a method that employs single-observer saliency maps has been already used for salien\-cy-aware video compression~\cite{SAVAM2}. The authors showed it could reduce the bitrate by 23\% compared with regular x264~\cite{x264} yet produce videos with the same visual quality.

\item Ground-truth saliency maps provide the ultimate quality, but their direct use in video compression cannot guarantee the best quality of compressed videos. These saliency maps are temporally inconsistent: the object of interest in a video could change quickly to another object. But video codecs rely heavily on temporal consistency and cannot effectively handle such fast changes.
\end{itemize}

\subsection{Saliency Models Preparation}
\label{subsec:models-preparation}

The prediction quality of deep neural network models improves considerably when they are trained on more data. This generalization problem is especially noticeable for saliency prediction in videos, as videos have more diversity than images but less training data. To create a practical saliency-aware video encoder, we studied how the generalization problem affects compressed-video quality.

In particular, we are interested in the extent to which compression quality declines when compressing a video with content that is unrepresented in the training set, as well as how fine-tuning could reduce this decline.

Our experiments used the SAVAM~\cite{Gitman2014} video-saliency dataset, which contains 45 videos that are each 12-18 seconds long.
The original models provided by the developers of OM-CNN, ACL and SAM were trained on others datasets: SAM used SALICON~\cite{SALICON}; ACL used SALICON, DHF1K~\cite{ACL-wang2018}, Hollywood-2~\cite{Hollywood2_UCFSports_Mathe2015} and UCF sports~\cite{Hollywood2_UCFSports_Mathe2015}; and OM-CNN used LEDOV~\cite{OM-CNN-Jiang2017}.
We were therefore able to employ SAVAM for the generalization problem.

We split the SAVAM dataset into training, testing and validation parts that consist of 23, 12 and 10 videos, respectively.
We fine-tuned the selected models using the training videos over 10 epochs.
The early stopping with a window size of three epochs is used to avoid overfitting.
Since ACL is a static model we fine-tuned it on every 25-th video frame.

The SAVAM dataset provides saliency information as the coordinates of fixation points for 50 observers. For fine-tuning and other experiments, we need ground-truth saliency maps in addition to fixation points. All our experiments convert fixation points to saliency maps using the formula $\textbf{SM}_p = \sum_{i=1..N}{\mathcal{N}(p, f_i, \sigma)}$, where $\textbf{SM}_p$ is the resulting saliency map value at pixel $p$, $f_i$ is the position of the $i$-th fixation point of $N$ and $\mathcal{N}$ is a Gaussian with $\sigma=120$.

Our initial model weights came from the authors of ACL and SAM-ResNet.
Since OM-CNN was trained only on the LEDOV video dataset, whereas ACL was trained on four different datasets, we for fairness initially retrained the OM-CNN model using the LEDOV, DIEM~\cite{DIEM_Mital2010}, Hollywood-2~\cite{Hollywood2_UCFSports_Mathe2015} and UCF~sports~\cite{Hollywood2_UCFSports_Mathe2015} video datasets then fine-tune it using SAVAM.
In Sections~\ref{sec:SMObjectiveEvaluation} and \ref{sec:CompressionObjectiveEvaluation} we compare the fine-tuned and original models.

The authors of~\cite{SAVAM2} showed that the performance of non-deep models can improve considerably through application of simple postprocessing to the saliency maps. They consider two postprocessing transformations---applying a brightness-correction function and blending with a precomputed center-prior image---and propose a method that computes their optimal parameters according to the MSE metric. We use this method and apply the postprocessing to saliency maps of the fine-tuned models and single-observer baseline.

We computed the parameters of the postprocessing transformations using the training part of the SAVAM dataset. The postprocessing can therefore be interpreted as an additional fine-tuning of the models, though it yields a noticeable gain only for non-deep models in general and the single-observer baseline in particular. We provide a detailed analysis of the fine-tuning and postprocessing in subsequent sections. Figure~\ref{fig:ModelExamples} shows example saliency maps after these steps.

\input{figures/saliency_objective_metrics.tex}

\subsection{Objective Saliency Models Evaluation}
\label{sec:SMObjectiveEvaluation}

We evaluated the baselines and three selected visual-attention models, along with their fine-tuned and postprocessed versions, using five objective-quality metrics for saliency~\cite{Bylinskii2016}: AUC-Judd (AUC-J), linear correlation coefficient (CC), Kullback-Leibler divergence (KL), normalized scanpath saliency (NSS) and similarity metric (SIM).
The results appear in Table~\ref{tab:SaliencyObjectiveMetrics}.

Table~\ref{tab:SaliencyObjectiveMetrics} shows that SAM-ResNet, ACL and OM-CNN exhibit higher quality after the fine-tuning: $0.517 \rightarrow 0.616$, $0.546 \rightarrow 0.604$ and $0.560 \rightarrow 0.569$, respectively, according to the SIM metric.
The other metrics demonstrate a notable improvement as well---except for AUC-J, where SAM-ResNet worsened after the fine-tuning.

Postprocessing boosted the single-observer baseline according to the SIM metric ($0.493 \rightarrow 0.603$) and the other metrics, except NSS ($2.465 \rightarrow 1.845$).

NSS is a good metric for evaluating saliency models~\cite{Bylinskii2016} because it uses fixation maps as ground-truth data instead of saliency maps, which cannot be unambiguously generated from fixation maps.
But NSS is poorly suited to evaluating model quality in the context of saliency-aware video compression. It encourages a model to generate saliency maps with sharp peaks next to fixation points, covering only a small part of the image, whereas the actual human gaze is fuzzier and covers wider parts of the image. Thus, the encoder will interpret many of the pixels that the observer sees around fixation points as nonsalient.

The postrocessing almost did not change the quality of the ACL and SAM-ResNet results (we saw a tiny boost in the CC, SIM and KL metrics but a drop in the others), and only the KL metric showed an improvement for OM-CNN.
The explanation could be that the models we considered have explicit center priors in their architectures and have already implicitly learned the postprocessing transformations.

All three selected models outperformed the center-prior baseline even without fine-tuning. The fine-tuned version of SAM-ResNet outperformed the postprocessed single-observer baseline, and whereas the fine-tuned ACL outperformed only the original single-observer model, all OM-CNN models are worse than the original single observer. Worth noting is that the results for the original SAM-ResNet were worse than those for the original ACL and OM-CNN all metrics, but after fine-tuning and postprocessing, it surpassed every ACL and OM-CNN version even though it is a purely static model whereas ACL and OM-CNN can exploit additional motion features.

\subsection{Objective Saliency-Aware Compression Evaluation}
\label{sec:CompressionObjectiveEvaluation}

\input{figures/compression_objective_metrics.tex}

Our model evaluation in the previous section showed that the fine-tuned SAM-ResNet exhibits the best objective quality for saliency maps. But the final quality of videos produced by a saliency-aware encoder depends on many factors, only one of which is the proximity of the saliency maps to ground-truth.

To evaluate the complete contribution of a saliency model to the final video quality, we compressed the validation videos at different bitrates using the method described in Section~\ref{sec:proposed} and measured the quality of the results using the EWSSIM~\cite{Li2011} metric.
EWSSIM is a simple measure of distortion between the compressed and original videos; it is the weighted sum of per-pixel SSIM values using ground-truth saliency as weights.

The resulting rate-distortion curves appear in Figure~\ref{fig:CompressionEWSSIM}. 
Note that for visibility we omitted some overlapping curves, but for each model we left at least one curve corresponding to a version with the best EWSSIM value.
As Figure~\ref{fig:CompressionEWSSIM} shows, the postprocessed single-observer baseline outperforms all saliency models.
In particular, it is slightly better than the postprocessed SAM-ResNet, whereas SAM-ResNet is the better of the two for all objective measures in Table~\ref{tab:SaliencyObjectiveMetrics}.
In general, the ranking of the other models by EWSSIM remains the same as the ranking by objective saliency measures.

SAM-ResNet despite of getting better score than the single-observer baseline according to AUC-J, CC, KL, NSS, and SIM, got lower  score accroding EWSSIM. The metric divergence could be explained by the fact that saliency objective metrics fail to take into account the temporal consistency of saliency maps, whereas this feature may be crucial for efficient video encoding. Since SAM-ResNet is a static model, it is less temporally stable than ACL and OM-CNN, which are designed for saliency prediction in video.

SAM-ResNet remains the best of our selected models according to EWSSIM, however. As Figure~\ref{fig:CompressionEWSSIM} shows, it can produce video at a 1000 kbps bitrate and the same EWSSIM quality as the original x264 at a 1330 kbps bitrate. Thus, the proposed saliency-aware compression with the SAM-ResNet model could reduce the bitrate by up to 25\%.

Also, the fine-tuning and postprocessing boosted the SAM-ResNet performance according to EWSSIM. The original SAM-ResNet yield\-ed only a 12\% bitrate savings. The fine-tuning effect for ACL and OM-CNN is more modest. The fine-tuned and postprocessed ACL yields a 22\% bitrate savings and the original version an 18\% savings.

But the EWSSIM metric cannot model all features of the human visual system. Therefore, we performed another subjective evaluation of the saliency-aware compression, as we describe in Section~\ref{sec:CompressionSubjectiveEvaluation}.

\subsection{Subjective Saliency-Aware Compression Evaluation}
\label{sec:CompressionSubjectiveEvaluation}

\input{figures/compression_subjective.tex}

To evaluate the performance of the selected saliency models for video compression, we conducted a study with over 301 paid participants using the  \href{http://www.subjectify.us/}{Subjectify.us} web service. 

Participants were shown a sequence of compressed-video pairs in a web browser using full-screen mode; for each pair, we asked them to either select the video with better quality or mark the videos as equal.

For this study, we used 10 validation videos from SAVAM dataset.
For each video we used our encoder to produce a set of compressed bitstreams by varying bitrate (1250, 1375 and 1500 kbps) and saliency prediction module: ground-truth, postprocessed single-observer and center-prior baselines as well as three deep models with fine-tuning and postprocessing. We also compressed each video with unmodified x264 at 1500 kbps bitrate.
Thus, we had 1710 video pairs that required comparison.
In total we collected 3010 pairwise comparisons from 301 participants,
distributed uniformly across all compression methods and test sequences.

In particular, each participant viewed 12 video pairs, two of which were hidden verification questions that asked them to compare regular x264 compression at bitrates of 1500 and 2500 kbps; our study only retained the results from those who correctly answered both verification questions.

We offered money reward to each participant who successfully completed the test. 
Using Thurstone's Case V Model~\cite{Thurstone}, we transformed the pairwise method comparisons into subjective rankings.
The final subjective ranks grouped by bitrate with 85\% confidence intervals are shown on Figure~\ref{fig:CompressionSubjective}.

As Figure~\ref{fig:CompressionSubjective} illustrates, the proposed saliency-aware compression at a 1500 kbps bitrate for all tested models ranks higher than regular x264 at the same bitrate.
This result means humans prefer saliency-aware compression (even with the simple center-prior model) to regular compression at equal bitrates.
At 1375 kbps, however, the subjective ranks shows that only SAM-ResNet certainly outperforms x264 at 1500 kbps; other models still rank close to x264, though.
At 1250 kbps, SAM-ResNet has almost the same rank as x264 at 1500 kbps, the other models (even Ground-Truth) have lower rank.
If we linearly interpolate model rankings, we can claim that saliency-aware compression using SAM-ResNet at birtate 1250 kbps has the same subjective quality as regular x264 at 1500 kbps, yielding a 17\% bitrate savings.

Note that by the subjective ranking SAM-ResNet beats the other models and the baselines (and even slightly better than Ground-Thruth at two bitrates), whereas the objective comparison by EWSSIM (see Section~\ref{sec:CompressionObjectiveEvaluation}) indicates that the postprocessed single-observer baseline achieves the best results.

\section{Conclusion}
\label{sec:conclusion}

In this paper we examined the problem of applying modern saliency models for saliency-aware video compression.
We proposed a saliency-aware video encoder with two modules.
The encoder module is based on x264 video encoder modification introduced in~\cite{SAVAM2}.
For the saliency prediction module we considered three deep saliency models and adapted them for video compression use case by applying special postprocressing and fine-tuning.
We objectively evaluated quality of original and adapted models as well as videos compressed using these models.
Paying special attention to quality evaluation of the resulting videos, we conducted a subjective experiment to estimate how humans perceive videos compressed using different models and bitrates.
The experiment showed that our method relying om SAM-ResNet for saliency prediction can produce videos with the same visual quality as unmodified x264, but with 17\% less bitrate.
Since the encoder module produces H.264 bitstream, our method can be applied to reduce video content delivery costs without need to upgrade the infrastructure and user devices to support new video encoding standards.


\bibliographystyle{ACM-Reference-Format}
\bibliography{bibliography.bib,new_compression.bib}

\end{document}

%% file: figures/encoder_schema.tex
\begin{figure}
    \centering
    \tikzstyle{block} = [draw, fill=white, rectangle, 
        minimum height=3em, minimum width=6em, text width=2cm, align=flush center]
    \tikzstyle{branch} = [fill, shape=circle, minimum size=3pt, inner sep=0pt]
    \tikzstyle{input} = [coordinate]
    \tikzstyle{output} = [coordinate]

    \begin{tikzpicture}[auto, node distance=2cm,>=latex']
        \node [input, name=input] {};
        \node [block, right=2.9cm of input] (encoder) {Encoding module};
        \path (input) -- coordinate (branch) (encoder);
        \node [block, below of=encoder] (saliency) {Saliency prediction module};
        \node [output, name=output, right=2.9cm of encoder] {};
        
        \draw[->] (branch) node[branch] {}{} |- (saliency);
        \draw [->] (input) -- node {Input video} (encoder);
        \draw [->] (saliency) -- (encoder);
        \draw [->] (encoder) -- node {Encoded stream} (output);
    \end{tikzpicture}

    \caption{Scheme of proposed saliency-aware encoder.}
    \label{fig:encoder-schema}
\end{figure}
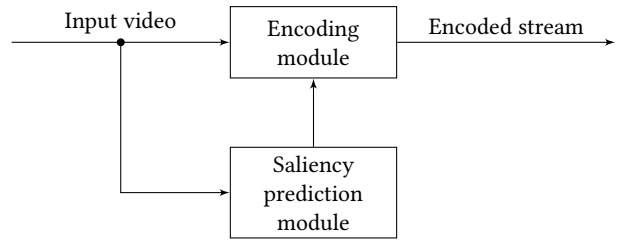

%% file: figures/models_examples.tex
\begin{figure*}[!th]
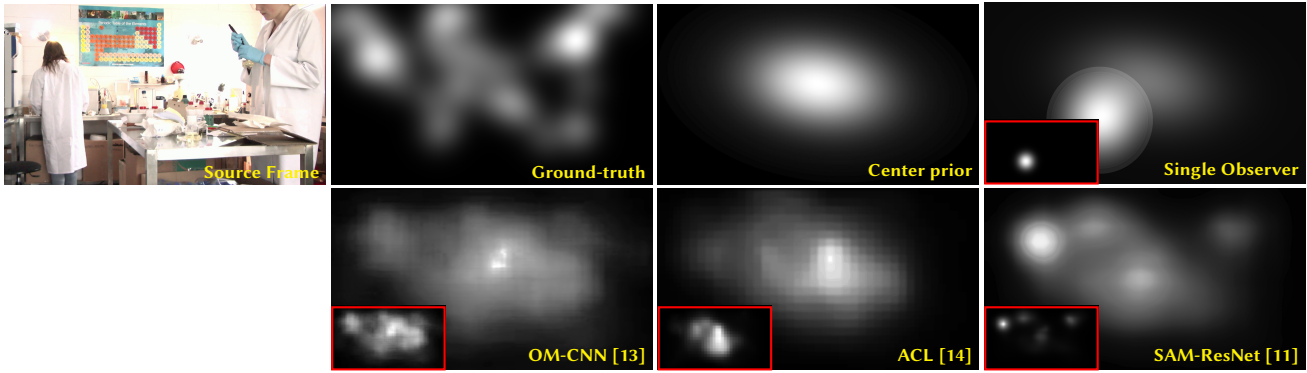

    \begin{center}\small\bfseries\sffamily

    	\newcommand{\contourblack}[1]{\textpdfrender{%
    			StrokeColor = black,
    			TextRenderingMode=FillStroke,
    			FillColor=yellow,
    			LineWidth=.07ex,
    		}{#1}
    	}

        \setlength{\tabcolsep}{1pt}
        \renewcommand{\arraystretch}{1}

        \begin{tabular}[h!]{p{0.24\linewidth}p{0.24\linewidth}p{0.24\linewidth}p{0.24\linewidth}}

            \begin{mtikzonimage}{\linewidth}{models_examples/frames.jpg}
                \node[color=yellow, left=2pt] at (\linewidth, 0.5\baselineskip) {\fontsize{7pt}{7pt}\selectfont Source Frame};
            \end{mtikzonimage}\stepcounter{mcounter} &

            \begin{mtikzonimage}{\linewidth}{models_examples/gt_saliency_rad=90_max_nrm.jpg}
                \node[color=yellow, left=2pt] at (\linewidth, 0.5\baselineskip) {\fontsize{7pt}{7pt}\selectfont Ground-truth};
            \end{mtikzonimage}\stepcounter{mcounter} &
            
            \begin{mtikzonimage}{\linewidth}{models_examples/CP.jpg}
                \node[color=yellow, left=2pt] at (\linewidth, 0.5\baselineskip) {\fontsize{7pt}{7pt}\selectfont Center prior};
            \end{mtikzonimage}\stepcounter{mcounter}&
        
            \begin{cropimg}{\linewidth}{models_examples/saliency_single-observer-postprocessed-gt-rad-120}
                      {0.35\linewidth}{models_examples/saliency_single-observer}
                \node[color=yellow, left=2pt] at (\linewidth, 0.5\baselineskip) {\fontsize{7pt}{7pt}\selectfont Single Observer};
            \end{cropimg}\stepcounter{mcounter}
            
        \tabularnewline[-0.4cm]
            
            &
        
            \begin{cropimg}{\linewidth}{models_examples/saliency_OM-CNN-4-datasets-postprocessed-gt-rad-120}
                     {0.35\linewidth}{models_examples/saliency_OM-CNN-4-datasets}
                \node[color=yellow, left=2pt] at (\linewidth, 0.5\baselineskip) {\fontsize{7pt}{7pt}\selectfont OM-CNN~\cite{OM-CNN-Jiang2017}};
            \end{cropimg}\stepcounter{mcounter} &

            \begin{cropimg}{\linewidth}{models_examples/saliency_acl_fine-tuned_best-loss-postprocessed-gt-rad-120}
                     {0.35\linewidth}{models_examples/saliency_acl_vanilla}
                \node[color=yellow, left=2pt] at (\linewidth, 0.5\baselineskip) {\fontsize{7pt}{7pt}\selectfont ACL~\cite{ACL-wang2018}};
            \end{cropimg}\stepcounter{mcounter} &
            
            \begin{cropimg}{\linewidth}{models_examples/saliency_sam-resnet_ft-cc-only-loss-best-postprocessed-gt-rad-120}
                     {0.35\linewidth}{models_examples/saliency_sam-resnet_original-resnet}
                \node[color=yellow, left=2pt] at (\linewidth, 0.5\baselineskip) {\fontsize{7pt}{7pt}\selectfont SAM-ResNet~\cite{SAM-cornia2018}};
            \end{cropimg}\stepcounter{mcounter}
            
        \end{tabular}
    \end{center}
    
    \vspace{-0.6cm}
    \caption{%
    	Saliency maps for three selected models and baselines.
    	Fine-tuned and postprocessed saliency maps of the models are shown.
    	Also, original saliency maps are shown in the left bottom corner of the image.
    }\label{fig:ModelExamples}
	
	\vspace{-0.45cm}
	
\end{figure*}

%% file: figures/saliency_objective_metrics.tex
\begin{table}[!t]

\centering
\resizebox{\linewidth}{!}{%
\begin{tabular}{lccccc}
\toprule
Model &  AUC-J$\uparrow$ &     CC$\uparrow$ &  KL$\downarrow$ &    NSS$\uparrow$ &  SIM$\uparrow$ \\
\midrule
\scriptsize Ground truth         &     -     &  -     &  -      &  3.250 & - \\
\scriptsize SAM-ResNet + FT + PP &     0.833 &  0.672 &   0.569 &  2.021 &             0.618 \\
\scriptsize SAM-ResNet + FT      &     0.835 &  0.664 &   1.242 &  2.180 &             0.616 \\
\scriptsize Single observer + PP &     0.807 &  0.651 &   0.621 &  1.845 &             0.603 \\
\scriptsize ACL + FT + PP        &     0.820 &  0.633 &   0.584 &  1.705 &             0.603 \\
\scriptsize ACL + FT             &     0.820 &  0.631 &   0.651 &  1.764 &             0.604 \\
\scriptsize Single observer      &     0.763 &  0.581 &   9.601 &  2.465 &             0.493 \\
\scriptsize OM-CNN + FT          &     0.791 &  0.560 &   0.863 &  1.504 &             0.569 \\
\scriptsize OM-CNN + FT + PP     &     0.790 &  0.557 &   0.691 &  1.417 &             0.563 \\
\scriptsize OM-CNN               &     0.781 &  0.553 &   1.652 &  1.462 &             0.560 \\
\scriptsize ACL                  &     0.796 &  0.552 &   1.985 &  1.769 &             0.546 \\
\scriptsize SAM-ResNet           &     0.842 &  0.543 &   2.818 &  2.220 &             0.517 \\
\scriptsize Center prior         &     0.751 &  0.469 &   1.731 &  1.087 &             0.503 \\
\bottomrule
\end{tabular}
}

\caption{Objective evaluation results of the selected models and the baselines using five metrics on the SAVAM~\cite{Gitman2014} dataset. The results include the original models, the fine-tuned models for the training part of the SAVAM dataset (FT), and the result after applying the postprocessing transformations (PP).}
\label{tab:SaliencyObjectiveMetrics}

\end{table}

%% file: figures/compression_objective_metrics.tex
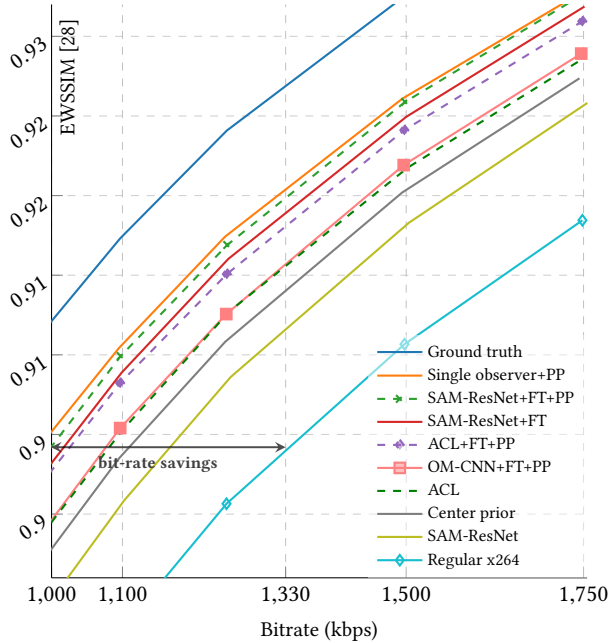
\begin{figure}[h]
\begin{tikzpicture}

\definecolor{color0}{rgb}{0.12156862745098,0.466666666666667,0.705882352941177}
\definecolor{color1}{rgb}{1,0.498039215686275,0.0549019607843137}
\definecolor{color2}{rgb}{0.172549019607843,0.627450980392157,0.172549019607843}
\definecolor{color3}{rgb}{0.83921568627451,0.152941176470588,0.156862745098039}
\definecolor{color4}{rgb}{0.580392156862745,0.403921568627451,0.741176470588235}
\definecolor{color5}{rgb}{0.549019607843137,0.337254901960784,0.294117647058824}
\definecolor{color6}{rgb}{0.890196078431372,0.466666666666667,0.76078431372549}
\definecolor{color7}{rgb}{0.737254901960784,0.741176470588235,0.133333333333333}
\definecolor{color8}{rgb}{0.0901960784313725,0.745098039215686,0.811764705882353}

\begin{axis}[
scale only axis,
width={0.84\linewidth},
height={0.9\linewidth},
xlabel={\small Bitrate (kbps)},
xtick={1000, 1100, 1330, 1500, 1750},
ylabel={\small EWSSIM~\cite{Li2011}},
legend cell align={left},
legend style={
anchor=south east, 
at={(1,0.0)}, 
fill=white,fill opacity=0.5, draw opacity=1,text opacity=1,
draw=none,
nodes={scale=0.82, transform shape},
font=\small
},
grid=both,
every minor grid/.style={dashed, opacity=0.00},
every major grid/.style={dashed, opacity=0.75},
y tick label style = {rotate=35, anchor=east},
tick align=inside,
tick pos=left,
x grid style={white!69.01960784313725!black},
xmin=1000, xmax=1755,
ymin=0.891, ymax=0.927,
axis x line*=bottom,
axis y line*=left,
y label style={anchor=north east, at={(0.0,0.99)}},
]

\addplot [thick, color0]
table [row sep=\\]{%
996.613569567247	0.906934140742808 \\
1096.89188826581	0.912319740077781 \\
1247.01723666354	0.919079196266946 \\
1497.8639347643	0.927530300254343 \\
1749.52700949242	0.933833279251177 \\
};
\addlegendentry{Ground truth};

\addplot [thick, color1]
table [row sep=\\]{%
995.145724225053	0.899910904018229 \\
1095.04121153692	0.905427256240007 \\
1244.73368484249	0.912424430013204 \\
1494.56155021397	0.921086290626344 \\
1745.36484663044	0.927706386432081 \\
};
\addlegendentry{Single observer+PP}

\addplot [thick, dashed, color2, mark=x]
table [row sep=\\]{%
997.290930644363	0.899081864653934 \\
1097.4388255509	0.904921205903883 \\
1247.6465733835	0.911897693619393 \\
1498.43804257664	0.920879816899288 \\
1749.54091682823	0.927642928275595 \\
};
\addlegendentry{SAM-ResNet+FT+PP}

\addplot [thick, color3]
table [row sep=\\]{%
998.802574437406	0.898137855316835 \\
1098.84298622313	0.903883091332698 \\
1249.20163685858	0.9110241159311 \\
1500.20880895481	0.919949659810052 \\
1751.36874047955	0.926871393266632 \\
};
\addlegendentry{SAM-ResNet+FT}

\addplot [thick, dashed, color4, mark=diamond*]
table [row sep=\\]{%
996.700184487191	0.89753970610526 \\
1096.69222705759	0.903231229632272 \\
1247.14227114829	0.910069593430288 \\
1497.3746408008	0.919109874252358 \\
1748.32376887058	0.925939892859649 \\
};
\addlegendentry{ACL+FT+PP}

\addplot [thick, mark=square*, color=red!50!white]
table [row sep=\\]{%
996.492411175105	0.894414148969645 \\
1096.73212613987	0.900401542336944 \\
1246.32314128101	0.907558917468283 \\
1496.45770794402	0.916914327666043 \\
1746.75977980999	0.923916549253798 \\
};
\addlegendentry{OM-CNN+FT+PP}

\addplot [thick, dashed, color=green!50!black]
table [row sep=\\]{%
999.675520133336	0.894475383242604 \\
1100.3032572018	0.900293765426863 \\
1250.79843172958	0.907730530846382 \\
1502.07384298405	0.916752109091441 \\
1753.49563135161	0.923715448401888 \\
};
\addlegendentry{ACL}

\addplot [thick, white!49.80392156862745!black, mark=o*]
table [row sep=\\]{%
995.135735585462	0.892521069187013 \\
1095.08931369057	0.898497394821583 \\
1244.90097214686	0.905822666893861 \\
1494.30170342638	0.915190558027724 \\
1744.31234040662	0.922364883514201 \\
};
\addlegendentry{Center prior}

\addplot [thick, color7]
table [row sep=\\]{%
1002.03627215803	0.889735127053242 \\
1102.37023042135	0.895856411074847 \\
1252.4882540838	0.90360489353176 \\
1503.7734038292	0.913292050008313 \\
1755.36383534785	0.920815970343256 \\
};
\addlegendentry{SAM-ResNet}

\addplot [thick, color8, mark=diamond]
table [row sep=\\]{%
995.950469740738	0.881268694671931 \\
1096.2280275556	0.887587612818464 \\
1246.67633442115	0.895650073224137 \\
1497.66370976475	0.905669321237843 \\
1748.45787297905	0.913449162362551 \\
};
\addlegendentry{Regular x264}

\fill[opacity=0.8, white] (1050.01,0.8992) rectangle (1250,0.8992);

\draw[stealth-stealth,line width=0.75pt, opacity=0.7] (999.269,0.8992) -- (1330,0.8992);

 \node at (1150,0.8992) [anchor=north, black!75!white] {\fontsize{6.5}{7}\selectfont \textbf{bit-rate savings}};

\end{axis}

\end{tikzpicture}

\vspace{-0.3cm}
\caption{Objective evaluation of proposed saliency-aware compression for different saliency models using the EWSSIM~\cite{Li2011} metric.}
\label{fig:CompressionEWSSIM}
\end{figure}

%% file: figures/compression_subjective.tex
\begin{figure}[th]

\begin{tikzpicture}
\begin{axis}[
width=\linewidth,
xlabel={\small Bitrate (kbps)},
ylabel={\small Subjective rank},
xtick={1250, 1375, 1500},
legend cell align={left},
legend style={
anchor=north west, 
at={(0.0,1.0)}, 
fill=white,fill opacity=0.25, draw opacity=1,text opacity=1,
draw=none,
font=\tiny
},
grid=both,
every minor grid/.style={thin, solid, opacity=0.0},
every major grid/.style={thin, solid, opacity=0.5},
tick align=inside,
tick pos=left,
x grid style={white!66!black},
xmin=1240, xmax=1500,
ymin=1.41, ymax=2.57,
axis x line*=bottom,
axis y line*=left,
]

\addplot [thick, green!80!black, mark=diamond*, error bars/.cd, y dir=both, y explicit]
plot coordinates {
(1250, 1.8138125049358316) +=(0, 0.191763109529423) -=(0, 0.19176310952942366)
(1375, 2.102174544075585) +=(0, 0.19653250192495397) -=(0, 0.19653250192495286)
(1500, 2.454652033084173) +=(0, 0.212023029985152) -=(0, 0.21202302998515288)
};
\addlegendentry{SAM-ResNet+FT+PP};

\addplot [thick, orange, mark=*]
plot coordinates {
(1250, 1.736000892290858)
(1375, 2.0726369782235747)
(1500, 2.507041849734532)
};
\addlegendentry{Ground truth};

\addplot [thick, gray, mark=square*]
plot coordinates {
(1250, 1.6402800678714964)
(1375, 1.7344220481531183)
(1500, 2.1120790676441503)
};
\addlegendentry{OM-CNN+FT+PP};

\addplot [thick, blue, mark=diamond]
plot coordinates {
(1250, 1.6062935534409908)
(1375, 1.6160258738768682)
(1500, 2.2757746390775813)
};
\addlegendentry{ACL+FT+PP};

\addplot [thick, pink!90!black, mark=*]
plot coordinates {
(1250, 1.4958027381197634)
(1375, 1.6885923412532682)
(1500, 2.0395038303481385)
};
\addlegendentry{Single observer+PP};

\addplot [thick, purple, mark=o]
plot coordinates {
(1250, 1.4509349190135097)
(1375, 1.5483096302371775)
(1500, 2.172323028595996)
};
\addlegendentry{Center prior};

\addplot [thick, dashed, red!80!black]
plot coordinates {
    (1000,1.788)
    (2000,1.788)
};
\addlegendentry{Regular x264, 1500 kbps};

\end{axis}
\end{tikzpicture}

\vspace{-0.3cm}
\caption{Subjective evaluation of the proposed saliency-aware compression using different saliency models. 85\% confidence intervals are shown for SAM-ResNet model.}
\label{fig:CompressionSubjective}
\end{figure}
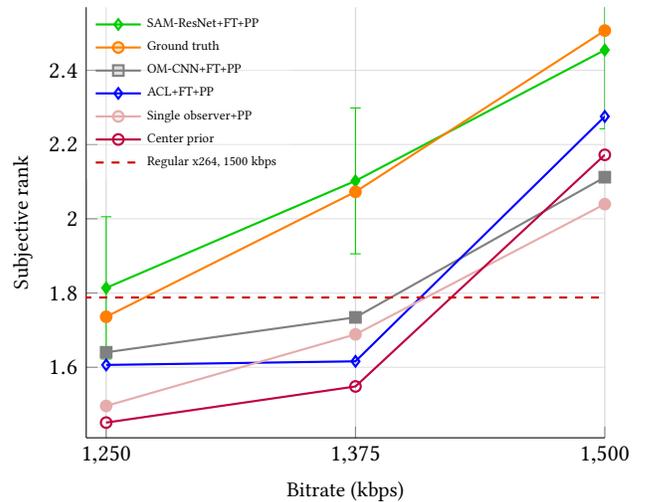

%% file: main.bbl
\begin{thebibliography}{10}

\bibitem{cisco-report}
Cisco.
\newblock Cisco visual networking index: Forecast and trends, 2017--2022.
\newblock Technical Report C11-741490-00, 2018.

\bibitem{Yarbus1967}
Alfred~L Yarbus.
\newblock {\em Eye movements during perception of complex objects}.
\newblock Springer, 1967.

\bibitem{SAVAM2}
Vitaliy Lyudvichenko, Mikhail Erofeev, Yury Gitman, and Dmitriy Vatolin.
\newblock A semiautomatic saliency model and its application to video
  compression.
\newblock In {\em 13th IEEE International Conference on Intelligent Computer
  Communication and Processing}, pages 403--410, 2017.

\bibitem{vp9-standard}
Adrian Grange, Peter de~Rivaz, and Jonathan Hunt.
\newblock Vp9 bitstream \& decoding process specification.
\newblock {\em Version 0.6, March}, 2016.

\bibitem{hevc-standard}
H~ITU-T.
\newblock 265: High efficiency video coding.
\newblock {\em ITU-T Recommendation H}, 265, 2013.

\bibitem{Borji2013}
A.~Borji and L.~Itti.
\newblock State-of-the-art in visual attention modeling.
\newblock {\em IEEE Transactions on Pattern Analysis and Machine Intelligence},
  35(1):185--207, Jan 2013.

\bibitem{Gitman2014}
Y.~Gitman, M.~Erofeev, D.~Vatolin, B.~Andrey, and F.~Alexey.
\newblock Semiautomatic visual-attention modeling and its application to video
  compression.
\newblock In {\em \ICIP}, pages 1105--1109, 2014.

\bibitem{Judd2012}
Tilke Judd, Fr{\'e}do Durand, and Antonio Torralba.
\newblock A benchmark of computational models of saliency to predict human
  fixations.
\newblock Technical report, Computer Science and Artificial Intelligence Lab,
  Massachusetts Institute of Technology, 2012.

\bibitem{Theis2015}
Lucas Theis and Matthias Bethge.
\newblock Deep gaze i: Boosting saliency prediction with feature maps trained
  on imagenet.
\newblock In {\em in ICLR Workshop}, 2015.

\bibitem{Kruthiventi2017}
S.~S.~S. Kruthiventi, K.~Ayush, and R.~V. Babu.
\newblock Deepfix: A fully convolutional neural network for predicting human
  eye fixations.
\newblock {\em IEEE Transactions on Image Processing}, 26(9):4446--4456, Sept
  2017.

\bibitem{SAM-cornia2018}
Marcella Cornia, Lorenzo Baraldi, Giuseppe Serra, and Rita Cucchiara.
\newblock {Predicting Human Eye Fixations via an LSTM-based Saliency Attentive
  Model}.
\newblock {\em IEEE Transactions on Image Processing}, 27(10):5142--5154, 2018.

\bibitem{Bak2018}
Cagdas Bak, Aysun Kocak, Erkut Erdem, and Aykut Erdem.
\newblock Spatio-temporal saliency networks for dynamic saliency prediction.
\newblock {\em IEEE Transactions on Multimedia}, 20(7):1688--1698, 2018.

\bibitem{OM-CNN-Jiang2017}
Lai Jiang, Mai Xu, and Zulin Wang.
\newblock Predicting video saliency with object-to-motion cnn and two-layer
  convolutional lstm.
\newblock {\em CoRR}, abs/1709.06316, 2017.

\bibitem{ACL-wang2018}
Wenguan Wang, Jianbing Shen, Fang Guo, Ming-Ming Cheng, and Ali Borji.
\newblock Revisiting video saliency: A large-scale benchmark and a new model.
\newblock 2018.

\bibitem{Borji2018}
Ali Borji.
\newblock Saliency prediction in the deep learning era: An empirical
  investigation, 2018.

\bibitem{Itti2004}
Laurent Itti.
\newblock Automatic foveation for video compression using a neurobiological
  model of visual attention.
\newblock {\em \TIP}, 2004.

\bibitem{polakovivc2018approach}
Adam Polakovi{\v{c}}, Radoslav Vargic, Gregor Rozinaj, and Gabriel-Miro
  Muntean.
\newblock An approach to video compression using saliency based foveation.
\newblock In {\em 2018 International Symposium ELMAR}, pages 169--172, 2018.

\bibitem{zund2013content}
Fabio Zund, Yael Pritch, Alexander Sorkine-Hornung, Stefan Mangold, and Thomas
  Gross.
\newblock Content-aware compression using saliency-driven image retargeting.
\newblock In {\em Image Processing (ICIP), 2013 20th IEEE International
  Conference on}, pages 1845--1849. IEEE, 2013.

\bibitem{gupta2011scheme}
Rupesh Gupta and Santanu Chaudhury.
\newblock A scheme for attentional video compression.
\newblock In {\em International Conference on Pattern Recognition and Machine
  Intelligence}, pages 458--465. Springer, 2011.

\bibitem{hadizadeh2014saliency}
Hadi Hadizadeh and Ivan~V Bajic.
\newblock Saliency-aware video compression.
\newblock {\em IEEE Transactions on Image Processing}, 23(1):19--33, 2014.

\bibitem{zhu2018spatiotemporal}
Shiping Zhu and Ziyao Xu.
\newblock Spatiotemporal visual saliency guided perceptual high efficiency
  video coding with neural network.
\newblock {\em Neurocomputing}, 2018.

\bibitem{sa-x264}
MSU~Video Group.
\newblock A fork of x264 video encoder supporting custom saliency maps as an
  additional input to improve quality of salient objects., 2017.

\bibitem{x264}
VideoLAN.
\newblock x264 software video encoder, 2004.

\bibitem{SALICON}
Xun Huang, Chengyao Shen, Xavier Boix, and Qi~Zhao.
\newblock Salicon: Reducing the semantic gap in saliency prediction by adapting
  deep neural networks.
\newblock In {\em 2015 International Conference on Computer Vision}, pages
  262--270, 2015.

\bibitem{Hollywood2_UCFSports_Mathe2015}
S.~Mathe and C.~Sminchisescu.
\newblock Actions in the eye: Dynamic gaze datasets and learnt saliency models
  for visual recognition.
\newblock {\em IEEE Transactions on Pattern Analysis and Machine Intelligence},
  pages 1408--1424, 2015.

\bibitem{DIEM_Mital2010}
Parag~K. Mital, Tim~J. Smith, Robin~L. Hill, and John~M. Henderson.
\newblock Clustering of gaze during dynamic scene viewing is predicted by
  motion.
\newblock {\em Cognitive Computation}, 3:5--24, 2010.

\bibitem{Bylinskii2016}
Zoya Bylinskii, Tilke Judd, Aude Oliva, Antonio Torralba, and Fr{\'{e}}do
  Durand.
\newblock What do different evaluation metrics tell us about saliency models?
\newblock {\em CoRR}, 2016.

\bibitem{Li2011}
Zhicheng Li, Shiyin Qin, and Laurent Itti.
\newblock Visual attention guided bit allocation in video compression.
\newblock {\em Image and Vision Computing}, 29(1):1--14, 2011.

\bibitem{Thurstone}
Louis~Leon Thurstone.
\newblock A law of comparative judgement.
\newblock {\em Psychological Review}, 34:278--286, 1927.

\end{thebibliography}
